\documentclass{egpubl}
\usepackage{egsr2025}
 
\SpecialIssuePaper         

\CGFStandardLicense

\usepackage[T1]{fontenc}
\usepackage{dfadobe}  

\usepackage{cite}  
\BibtexOrBiblatex
\electronicVersion
\PrintedOrElectronic
\ifpdf \usepackage[pdftex]{graphicx} \pdfcompresslevel=9
\else \usepackage[dvips]{graphicx} \fi

\usepackage{egweblnk} 

\usepackage[cjk]{kotex}
\usepackage{lipsum}
\usepackage{amssymb}
\usepackage{multirow}
\usepackage{xcolor}
\usepackage{colortbl}
\usepackage{booktabs}  
\usepackage{amsmath}
\usepackage[utf8]{inputenc}
\newcommand{\Eq}[1]  {Eq.\ (\ref{eq:#1})}

\newcommand{\Fig}[1] {Figure \ref{fig:#1}}
\newcommand{\Figs}[1]{Figures \ref{fig:#1}}
\newcommand{\Tbl}[1]  {Table \ref{tbl:#1}}
\newcommand{\Tbls}[1] {Tables \ref{tbl:#1}}
\newcommand{\Sec}[1] {Section \ref{sec:#1}}

\definecolor{silver}{rgb}{0.75, 0.75, 0.75}
\definecolor{golden}{rgb}{1.0, 0.87, 0.0}
\definecolor{copper}{rgb}{0.72, 0.45, 0.2}
\definecolor{bronze}{rgb}{0.8, 0.5, 0.2}

\makeatletter
\def\RemoveSpaces#1{\zap@space#1 \@empty}
\makeatother

\title
{Multiview Geometric Regularization of Gaussian Splatting for Accurate Radiance Fields}

\author[J. Kim, G. Park, S. Lee]
{\parbox{\textwidth}{\centering Jungeon Kim\orcid{0000-0003-4212-1970},
        Geonsoo Park\orcid{0009-0007-3839-3582},
        and Seungyong Lee\orcid{0000-0002-8159-4271} 
        }
        \\
{\parbox{\textwidth}{\centering POSTECH, South Korea
       }
}
}

\begin{document}
\teaser{
 \includegraphics[width=0.9\linewidth]{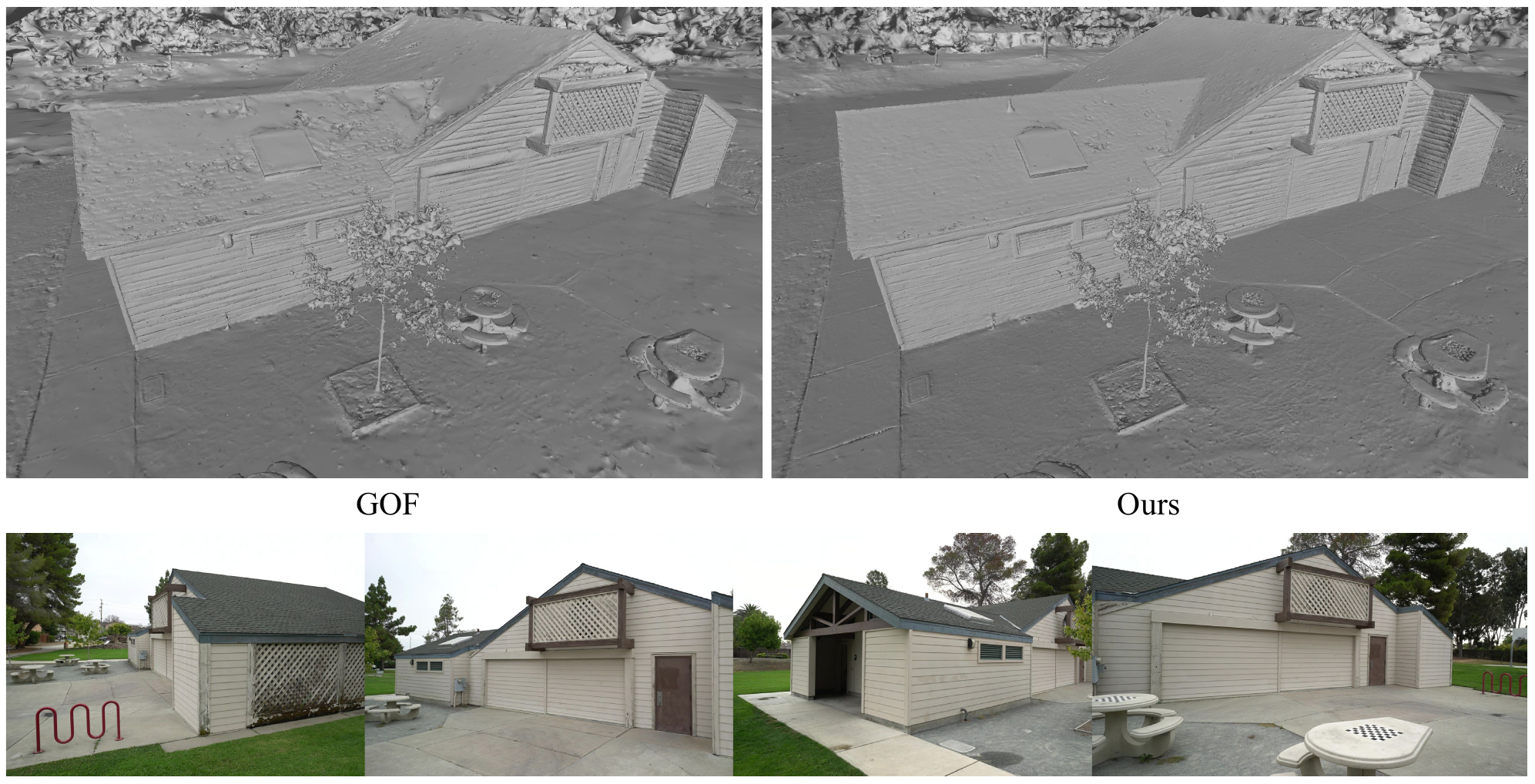}
 \centering
  \caption{ Comparison of our and GOF~\cite{yu2024gof} reconstructions from RGB images on the Tanks and Temple `Barn' scene~\cite{knapitsch2017tanks}. Our method reconstructs high-fidelity and smooth surfaces, compared to the state-of-the-art Gaussian Splatting-based surface reconstruction approach~\cite{yu2024gof}. The bottom row shows some of the input RGB images.
  }
\label{fig:teaser}
}

\maketitle
\begin{abstract}
Recent methods, such as 2D Gaussian Splatting and Gaussian Opacity Fields, have aimed to address the geometric inaccuracies of 3D Gaussian Splatting while retaining its superior rendering quality. However, these approaches still struggle to reconstruct smooth and reliable geometry, particularly in scenes with significant color variation across viewpoints, due to their per-point appearance modeling and single-view optimization constraints. In this paper, we propose an effective multiview geometric regularization strategy that integrates multiview stereo (MVS) depth, RGB, and normal constraints into Gaussian Splatting initialization and optimization. Our key insight is the complementary relationship between MVS-derived depth points and Gaussian Splatting-optimized positions: MVS robustly estimates geometry in regions of high color variation through local patch-based matching and epipolar constraints, whereas Gaussian Splatting provides more reliable and less noisy depth estimates near object boundaries and regions with lower color variation. To leverage this insight, we introduce a median depth-based multiview relative depth loss with uncertainty estimation, effectively integrating MVS depth information into Gaussian Splatting optimization. We also propose an MVS-guided Gaussian Splatting initialization to avoid Gaussians falling into suboptimal positions. Extensive experiments validate that our approach successfully combines these strengths, enhancing both geometric accuracy and rendering quality across diverse indoor and outdoor scenes.

\ccsdesc[300]{Computing methodologies~Point-based models}

\printccsdesc   
\end{abstract}  
\section{Introduction}
\label{sec:intro}

Recent advances in neural rendering have yielded impressive capabilities for novel-view synthesis from posed RGB images. Among these, 3D Gaussian Splatting (3DGS)~\cite{kerbl20233dgs} has garnered significant attention due to its remarkable ability to reconstruct radiance fields rapidly and render high-quality novel views in real-time. Its computational efficiency and rendering quality have spurred numerous downstream research efforts aimed at further refining and extending its capabilities~\cite{xu2024texture,yu2024mipsplatting,wang2024learning}. 3DGS represents scenes using a collection of 3D Gaussians, optimizing their parameters (position, shape, opacity, and view-dependent color) through differentiable rendering. However, the 3DGS formulation lacks explicit mechanisms to enforce multiview geometric consistency. Consequently, 3D surfaces extracted from optimized Gaussian parameters using standard techniques like Poisson surface reconstruction~\cite{kazhdan2013screened} or truncated signed distance function (TSDF) fusion~\cite{curless1996volumetric} often suffer from significant inaccuracies, noise, and a lack of detail~\cite{guedon2024sugar, huang20242dgs}.

To mitigate these geometric shortcomings, several methods enhancing Gaussian splatting have recently emerged~\cite{guedon2024sugar, huang20242dgs,yu2024gof}. SuGaR~\cite{guedon2024sugar} incorporates signed distance-induced regularization, encouraging Gaussians to align with an underlying surface. 2D Gaussian Splatting (2DGS)~\cite{huang20242dgs} introduces a 2D Gaussian representation in the form of disks, which naturally approximate local surface patches as planes and enable multiview consistent rendering. It also proposes single-view regularization terms for normal consistency and depth distortion to promote noise-free and smoother surfaces. Gaussian Opacity Fields (GOF)~\cite{yu2024gof} formulate the contribution of a Gaussian to a ray with explicit ray-Gaussian intersection, enabling the construction of an opacity field, and also utilize the two regularization terms from 2DGS. Despite these advances, approaches relying primarily on such single-view-based geometric regularization methods or implicit surface constraints remain limited in their ability to capture robust and accurate multiview consistent geometry across diverse scenes.

Parallel to these rendering-focused methods, multiview stereo (MVS) has long been a fundamental technique for accurate 3D geometry reconstruction from posed images~\cite{schoenberger2016mvs,chen2024multiview}. Conventional MVS algorithms typically employ patch-based matching between corresponding image regions, enforcing epipolar constraints, to estimate dense depth maps. These methods excel at producing accurate geometry in well-textured areas and demonstrate resilience to moderate lighting variations. However, they frequently struggle near object boundaries and in texture-less regions, resulting in noisy or incomplete depth estimates~\cite{chen2024multiview}.

\begin{figure}[t]
  \centering
  
  \includegraphics[width=\linewidth]{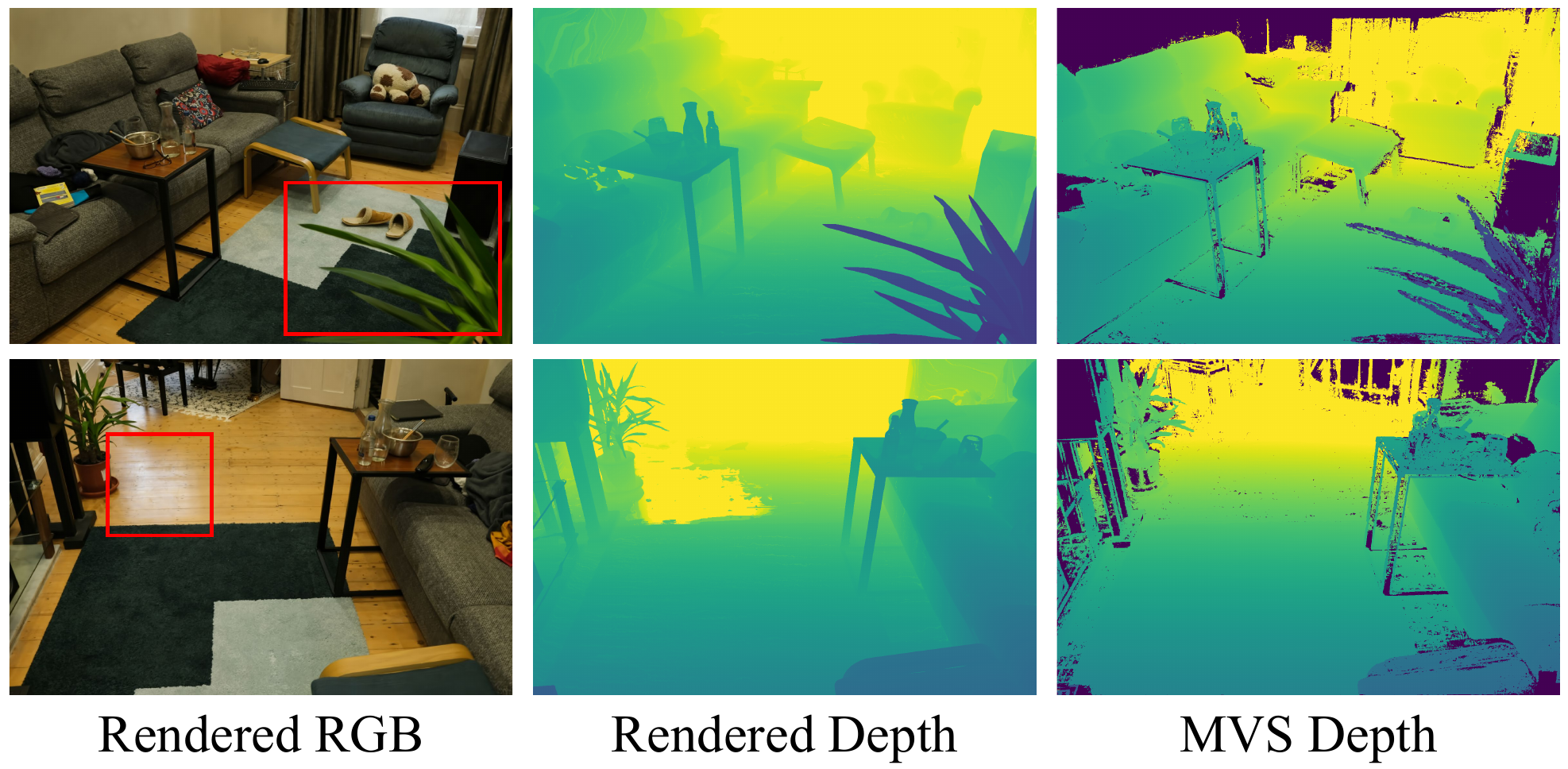}
 
  \caption{\label{fig:mvs_vs_GS}
          Complementary relationship between depths estimated by Multiview Stereo (MVS)~\cite{schoenberger2016mvs} and depths rendered by GOF~\cite{yu2024gof}. MVS estimates accurate depths in well-textured regions but often produces noisy depths near object boundaries (top). In contrast, GOF effectively represents sharp object boundaries but may yield geometrically inaccurate surfaces, particularly in regions with view-dependent appearance variations such as lighting and specularity.
           }
\end{figure}

A key observation motivating our work is the complementary nature of MVS-derived geometry and Gaussian splatting reconstructions. MVS provides precise 3D points in regions rich with visual features but often yields noisy estimates near depth discontinuities or homogeneous areas. Conversely, 3DGS, with its per-point appearance modeling, adeptly captures complex view-dependent effects and can represent sharp object boundaries through the learned opacity and spatial distribution of its Gaussians. However, this very flexibility, particularly the per-point view-dependent appearance, can hinder the optimization process from converging to a geometrically consistent surface across all viewpoints. This challenge is especially pronounced in real-world scenes exhibiting significant appearance variations (e.g., due to lighting or specularity), where the optimization might prioritize fitting appearance over achieving geometric accuracy (\Fig{mvs_vs_GS}).

Leveraging this complementarity, we propose an effective \textit{multiview geometric regularization} strategy for Gaussian Splatting that integrates the strengths of both MVS and 3DGS. Our goal is to achieve radiance fields that are accurate in both rendered appearance and underlying geometry. Crucially, our regularization is applied at both the initialization and optimization stages of the Gaussian splatting pipeline.

During optimization, we utilize MVS depth priors to guide the Gaussians towards the underlying scene surface. To this end, we introduce a \textit{median-depth-based multiview relative depth loss} incorporating uncertainty from the rendering process itself. For each pixel ray, we compute the rendered median depth via accumulated transmittance thresholding, representing the current estimated surface location, and crucially, we also estimate the uncertainty of this rendered median depth based on the total accumulated transmittance of the ray. Our loss then encourages the positions of Gaussians contributing significantly near this rendered location to align with corresponding MVS depth estimates across relevant views. The strength of this alignment guidance is modulated by the certainty of the rendered median depth.

Importantly, relying solely on geometric loss terms during optimization often struggles with suboptimal local minima when starting from sparse or inaccurate Gaussian positions. To mitigate this, we propose a robust MVS-guided initialization procedure. By utilizing geometric information derived from MVS~\cite{schoenberger2016mvs}, we establish a strong, geometrically-aware initial state for the Gaussians, significantly improving the final reconstruction quality (\Fig{ablation} (f)).

Furthermore, to promote multiview consistent appearance, we incorporate multiview RGB photometric losses. We also extend the normal consistency and depth distortion regularizers, previously used in single-view contexts~\cite{huang20242dgs}, into multiview formulations that explicitly enforce geometric smoothness and agreement across multiple viewpoints. These geometric regularizers play a crucial role especially in homogeneous areas where the MVS depth prior may be noisy or missing, thus complementing the MVS guidance.

Our contributions include: (1) An MVS-guided initialization strategy tailored for robust Gaussian splatting optimization; (2) A median-depth-based multiview relative depth loss incorporating uncertainty for optimization-time geometric regularization; and (3) Multiview extensions of geometric consistency losses for faithful and smooth geometry reconstruction. Through extensive experiments on diverse and challenging datasets, including Mip-NeRF360~\cite{barron2022mipnerf}, DTU~\cite{jensen2014large}, and Tanks and Temples~\cite{knapitsch2017tanks}, we demonstrate that our multiview geometric regularization strategy is highly effective. Our evaluations show that our approach outperforms existing state-of-the-art models in both novel-view synthesis quality and, critically, 3D surface reconstruction accuracy.

\section{Related Work}

\subsection{Novel View Synthesis}
\paragraph*{Conventional Methods}

Traditional novel view synthesis often relied on explicit 3D reconstruction methods~\cite{seitz2006mvs, snavely2006photo, goesele2007multi, schonberger2016sfm}. These methods struggled with complex geometries, non-Lambertian materials, and realistic rendering of complex visual effects, including reflections and transparency, often requiring dense views.

\paragraph*{Neural Radience Field}
Mildenhall et al.~\cite{mildenhall2021nerf} first introduced Neural Radiance Fields (NeRF), utilizing neural networks such as multi-layer perceptrons (MLPs) to directly learn a continuous volumetric scene representation from a limited set of input views. The implicit scene representation achieved state-of-the-art results in synthesizing photorealistic novel views of complex scenes that conventional methods often fail to represent. 
Inspired by these promising results, subsequent studies have extensively improved NeRF by addressing key limitations in training speed, efficiency, and rendering quality. Mip-NeRF360~\cite{barron2022mipnerf} effectively handled unbounded scenes and rendering aliasing, improving rendering quality but modestly increasing training time.  Instant-NGP~\cite{muller2022instant} drastically improved training and rendering speed using multi-resolution hash encodings. Others focused on compression (MERF~\cite{reiser2023merf}) or converting implicit fields to explicit ones for faster rendering (SNeRG~\cite{hedman2021snerg}, BakedSDF~\cite{yariv2023bakedsdf}). Despite advances, NeRF-based methods often have high computational costs, motivating alternatives like 3DGS.

\paragraph*{Gaussian Splatting}
Kerbl et al.~\cite{kerbl20233dgs} recently introduced 3D Gaussian Splatting (3DGS) for fast training and real-time rendering of radiance fields. This technique represents scenes as a collection of millions of 3D anisotropic Gaussians, which are rendered using a splatting-based rasterizer. Numerous follow-up works have emerged to address the limitations of 3DGS and enhance its performance. For instance, Mip-Splatting~\cite{yu2024mipsplatting} addressed aliasing and frequency artifacts by introducing 3D smoothing filters and a mipmap-style anti-aliasing. Other research line, such as Compressed-3DGS~
\cite{niedermayr2024compressed} and LightGaussian~\cite{fan2024lightgaussian} has focused on reducing the excessive number of Gaussians to make the method suitable for network streaming and low-powered mobile devices.
Scene representation has also been an active area of research, with proposals for variants like 2D Gaussians~\cite{dai2024gaussiansurfels,huang20242dgs} and hybrid representations~\cite{yu2024gsdf,lu2024scaffold}. 
Furthermore, generalizable Gaussian Splatting models have recently been investigated~\cite{charatan2024pixelsplat,chen2024mvsplat}. These models can directly predict Gaussian Splatting parameters for radiance field representation without the need for per-scene optimization. However, such methods are primarily designed for sparse-view settings and do not achieve the high-fidelity rendering demonstrated by per-scene optimization techniques. Our method falls into the per-scene optimization category and aims for high performance. To achieve this, we leverage multiview stereo (MVS) depths for effective geometric regularization within the Gaussian Splatting optimization process.

The utilization of depth priors is not a new concept. Before the advent of 3DGS, point-based rendering methods commonly used 3D points derived from MVS depth images as proxy geometry~\cite{kopanas2021point,kopanas2022neural,riegler2021stable,riegler2020free}. 
However, 3DGS has shown superior rendering quality and efficiency over these point-based methods. 
Recently, a few works have revisited the use of MVS depth priors to improve the robustness of 3DGS in sparse-view settings~\cite{wu2025sparse2dgs,shen2024solidgs}. 
Concurrent with our work, Li et al.~\cite{li2025mpgs} utilize monocular depth priors, instead of MVS depth, to better handle weakly textured regions in 3DGS.

\subsection{Neural Surface Reconstruction}
Neural surface reconstruction approaches are largely divided into two categories: reconstruction from point clouds and reconstruction from multi-view images. This section focuses on methods based on multi-view images, as they are most relevant to our work, rather than those based on point clouds~\cite{ma2020neural, ma2022surface, erler2020points2surf}.

\paragraph*{Implicit Surface Representation}

Implicit surface reconstruction methods represent geometry as continuous fields without relying on explicit primitives like meshes. Methods such as NeuS~\cite{wang2021neus} and VolSDF~\cite{yariv2021volsdf} leveraged neural implicit representations based on signed distance functions to reconstruct detailed surface geometries. Wang et al.~\cite{wang2023neus2} significantly accelerated neural implicit surface reconstruction using multi-resolution hash encodings and CUDA parallelization. Oechsle et al.~\cite{oechsle2021unisurf} unified implicit surface models with neural radiance fields, enabling accurate reconstruction without masks through combined volume and surface rendering. Fu et al.~\cite{fu2022geoneus} explicitly imposed multi-view geometric constraints on neural implicit surfaces, significantly improving geometry consistency and reconstruction quality for both thin structures and smooth regions.
Neuralangelo~\cite{li2023neuralangelo} leverages multi-resolution 3D hash grids with a progressive, coarse-to-fine optimization strategy and numerical gradient computation to reconstruct highly detailed and large-scale surfaces.
Despite their accuracy, these implicit methods often require extensive optimization times. 

\paragraph*{Explicit Surface Representation}

Explicit surface reconstruction methods directly define and optimize geometric primitives. Several recent approaches adapt 3D Gaussian Splatting for this purpose by encouraging the Gaussians to conform to an underlying surface. For example, SuGaR~\cite{guedon2024sugar} incorporates a signed distance-induced regularization to promote this alignment.
Other methods modify the Gaussians themselves for better geometric representation. 2DGS~\cite{huang20242dgs} and GaussianSurfel~\cite{dai2024gaussiansurfels} flatten 3D Gaussians into 2D counterparts, often with additional regularization terms to enhance geometric fidelity. To handle large-scale, unbounded scenes, GOF~\cite{yu2024gof} constructs a Gaussian opacity field on a tetrahedral grid, enabling efficient and high-quality mesh extraction.
Building on these explicit frameworks, we propose a novel regularization strategy that leverages MVS depth priors to further advance geometric accuracy. This approach significantly improves both geometry and rendering quality, particularly in challenging scenes with substantial view-dependent color variations.

\begin{figure*}[tp]
  \centering
 
  \includegraphics[width=.99\linewidth]{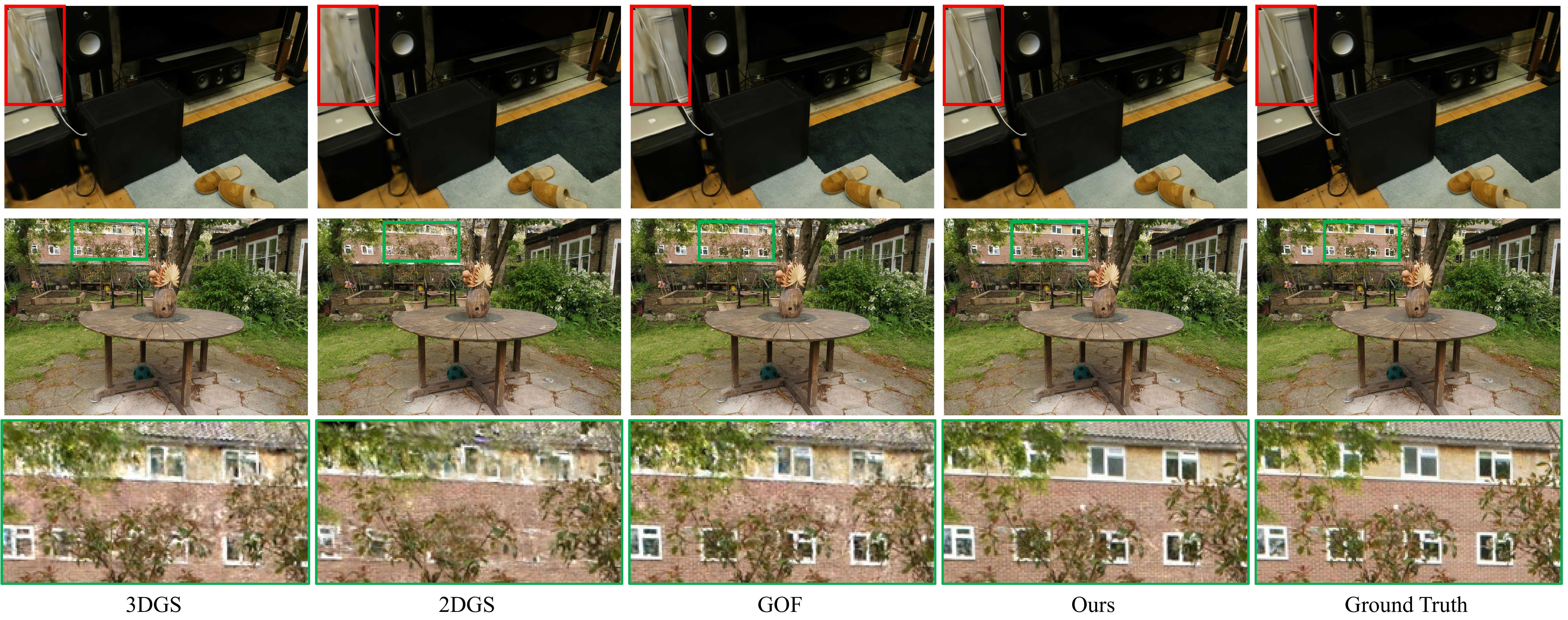}
  
  \caption{\label{fig:comp_nerf360_nvs} Visual comparison of novel view synthesis quality on the Mip-NeRF360 'room' and 'garden' scenes~\cite{barron2022mipnerf}. Our method faithfully reconstructs the appearance of scenes, compared to other state-of-the-art approaches (3DGS~\cite{kerbl20233dgs}, 2DGS~\cite{huang20242dgs}, GOF~\cite{yu2024gof}).}
\end{figure*}

\section{Overview}
\label{sec:overview}
The goal of our framework is to reconstruct high-fidelity, geometrically accurate radiance fields represented by 3D Gaussians, given a set of multiview RGB images with corresponding camera poses, typically estimated by Structure-from-Motion (SfM)~\cite{schonberger2016sfm}. As a preprocessing step, we first estimate dense depth maps for input RGB images using a conventional PatchMatch-based multiview stereo (MVS) method and then filter unreliable initial depths using heuristics based on multiview geometric consistency~\cite{schoenberger2016mvs}. We also determine triplets of adjacent viewpoints by leveraging image feature correspondences from SfM, which are utilized during our multiview optimization stage. Our framework builds upon the representation of Gaussian Opacity Fields (GOF)~\cite{yu2024gof} and its differentiable rendering formulation for Gaussian parameter optimization. 
Our multiview geometric regularization strategy plays a key role in both Gaussian parameter initialization and optimization to achieve accurate and smooth geometry reconstruction while maintaining high-quality appearance modeling.

Finally, the optimized Gaussian parameters are used to render high-quality novel views and can also produce median depth maps or opacity fields. These outputs can then be utilized with techniques like Truncated Signed Distance Function (TSDF) fusion~\cite{curless1996volumetric} and Marching Tetrahedra~\cite{yu2024gof} to extract a high-quality mesh. We first briefly review the GOF model in \Sec{gs_modeling} and then elaborate on our proposed multiview geometric regularization strategy in \Sec{gs_mv_reg}.

\section{Gaussian Splatting Model}
\label{sec:gs_modeling}

\paragraph*{Representation}

Gaussian Opacity Fields (GOF)~\cite{yu2024gof} builds upon the framework of 3D Gaussian Splatting (3DGS)~\cite{kerbl20233dgs} by explicitly incorporating ray-tracing-based volume rendering. Like 3DGS, the scene is represented by a collection of 3D Gaussians $\{\mathcal{G}_k\}$, where the parameters of a 3D Gaussian consist of a center $\mathbf{p}_k \in \mathbb{R}^3$, a scaling matrix $\mathbf{S}_k \in \mathbb{R}^{3\times3}$,
and a rotation $ \mathbf{R}_k \in \mathbb{R}^{3\times3}$.
The key difference lies in the methodology for evaluating the contribution of a Gaussian to pixels in the volume rendering equation.
While 3DGS calculates the contribution of a 3D Gaussian onto a pixel via a 2D Gaussian by projecting the 3D Gaussian onto the image, GOF directly computes the contribution of a 3D Gaussian for a pixel ray by analytically finding the maximum of the Gaussian values evaluated along the ray. 
Consequently, the contribution $\mathcal{E}(\mathcal{G}_k, \mathbf{o}, \mathbf{r})$ for a pixel ray $\mathbf{o}+t\mathbf{r}$ is incorporated into the volume rendering equation as follows:

\begin{align}
    \mathbf{c}(\mathbf{o}, \mathbf{r}) &= \sum_{k=1}^{K} \mathbf{c}_k \, \alpha_k \, \mathcal{E}(\mathcal{G}_k, \mathbf{o}, \mathbf{r}) T_k, \\ 
    T_k&= \prod_{j=1}^{k-1}(1 - \alpha_j \mathcal{E}(\mathcal{G}_j, \mathbf{o},  \mathbf{r})) \nonumber
\end{align}

\noindent
where
$\alpha_k \in [0,1]$ modulates Gaussian opacity globally, $\mathbf{c}_k$ is the view-dependent color modeled via spherical harmonics, and $T_k$ is the accumulated transmittance.

\paragraph*{Loss} Like 3DGS,
GOF optimization initiates from a sparse SfM point cloud and minimizes the following objective:

\begin{gather} 
\mathcal{L} = \mathcal{L}_c + \lambda_d\mathcal{L}_d + \lambda_n\mathcal{L}_n,
\end{gather}

\noindent where $\mathcal{L}_c$ is the RGB reconstruction loss~\cite{kerbl20233dgs}, a depth distortion loss $\mathcal{L}_d$, and a normal consistency loss $\mathcal{L}_n$ serve as geometric regularization terms~\cite{huang20242dgs}. The depth distortion loss is defined as $\mathcal{L}_d = \sum_{i,j}\omega_i\omega_j|t_i - t_j|$, where indices $i,j$ run over Gaussians contributing to a pixel ray $\mathbf{o}+t\mathbf{r}$, and $\omega_i = \alpha_i\mathcal{E}(\mathcal{G}_i,\mathbf{o},\mathbf{r})T_i$ is the blending weight of the $i$-th Gaussian. Here, $t_i$ denotes the intersection depth of $\mathcal{G}_i$ with the pixel ray. The normal consistency loss aligns the gradient of the rendered depth with the normal of the 3D Gaussian and is defined as $\mathcal{L}_n = \sum_{i} \omega_i (1 - \mathbf{n}_i^\top \mathbf{N})$, where the index $i$ runs over Gaussians intersected along the ray, $\mathbf{N}$ is the normal estimated from the gradient of the rendered depth~\cite{huang20242dgs}, and $\mathbf{n}_i$ is the normal vector at the ray-Gaussian intersection plane. For further details, refer to the GOF paper~\cite{yu2024gof}.

\section{Geometric Regulation of Gaussian Splatting Optimization}
\label{sec:gs_mv_reg}

\subsection{Geometric Regularization during Optimization}
\paragraph*{Relative Depth Loss}

As discussed in \Sec{intro}, the geometry reconstructed by Multiview Stereo (MVS) and Gaussian Splatting are complementary (\Fig{mvs_vs_GS}). However, MVS depth estimates often remain noisy even after applying geometric constraint-based heuristic filtering. Consequently, MVS depths cannot be reliably treated as ground truth due to residual noise. This necessitates methods to effectively handle these potentially inaccurate MVS priors.

Our observations indicate that while optimized Gaussian positions may not be precisely aligned with the underlying surface, they are typically distributed in proximity to it. Based on this premise, we utilize the rendered depth as a reference to facilitate the identification and rejection (via thresholding) of potentially erroneous MVS depth values. Specifically, our single-view relative depth loss, which incorporates the MVS depth prior, is defined as:

\begin{gather} \label{eq:relative_depth_loss}
\mathcal{L}_{rel} = \left|\,1 - \frac{\mathbf{D}_{mvs}}{\mathbf{D}_r}\,\right| \cdot \mathbf{U} \cdot \mathbb{I}(|\mathbf{D}_r - \mathbf{D}_{mvs}| < s \cdot \mathbf{D}_r),
\end{gather}

\noindent where $\mathbf{D}_{mvs}$ is the MVS depth, $\mathbf{D}_r$ is the rendered depth, $s$ is a hyperparameter for the thresholding tolerance (which is typically annealed from a less restrictive to a more restrictive value during optimization), $\mathbb{I}$ is
the indicator function yielding 1 if the condition holds and 0 otherwise, and $\mathbf{U}$ represents the rendering certainty, defined as the accumulated alpha $\mathbf{U}=\Sigma_i \omega_i$. 
Owing to its relative formulation, this loss exhibits high sensitivity to errors at closer depths.

For computing the rendered depth $\mathbf{D}_r$, a straightforward approach is to calculate the mean depth using the volume rendering weights $\omega_i$  and the depths $t_i$ corresponding to Gaussian contributions. However, the mean depth computation can lead to inaccurate estimates, particularly for semi-transparent or transparent objects. Even with opaque objects, the mean depth necessarily introduces artifacts analogous to "flying pixels" near object boundaries. Furthermore, reliance on the mean depth, calculated using potentially unstable weights $\omega_i$
during optimization, can harm reconstruction quality (\Fig{ablation} (d)). Therefore, we adopt the median depth, which effectively mitigates the aforementioned difficulties. Following~\cite{huang20242dgs}, we compute the median depth as the largest depth value t considered visible, employing $T_i>0.5$ as the threshold differentiating the surface from free space. Its definition is $D_r = \max \{ t_i \mid T_i > 0.5 \}$

\paragraph*{Multiview Geometric Regularization}
While enforcing the median-depth-based relative depth loss (\Eq{relative_depth_loss}) at a single viewpoint during optimization substantially improves geometric fidelity, this single-view application may not effectively regularize all scene regions. This limitation can arise because the optimization gradient primarily affects the parameters of Gaussians contributing to the median depth calculation in that specific view. To address this limitation, we extend the single-view loss \Eq{relative_depth_loss} to a multiview formulation, thereby enabling parameter updates informed by information from multiple viewpoints simultaneously within each optimization iteration. For similar reasons, we extend the normal consistency ($\mathcal{L}_n$) and depth distortion ($\mathcal{L}_d$) losses (defined in \Sec{gs_modeling}) to multiview versions. Finally, to promote multiview consistent appearance, we incorporate multiview RGB photometric losses ($\mathcal{L}_c$). Consequently, our final objective function is formulated as:

\begin{equation}
\mathcal{L} = \sum_v (\mathcal{L}^v_c + \lambda_{rel} \mathcal{L}^v_{rel} + \lambda_d \mathcal{L}^v_d + \lambda_n \mathcal{L}^v_n)
\end{equation}

\subsection{MVS-guided initialization}

\begin{figure*}[th]
  \centering

  \includegraphics[width=.95\linewidth]{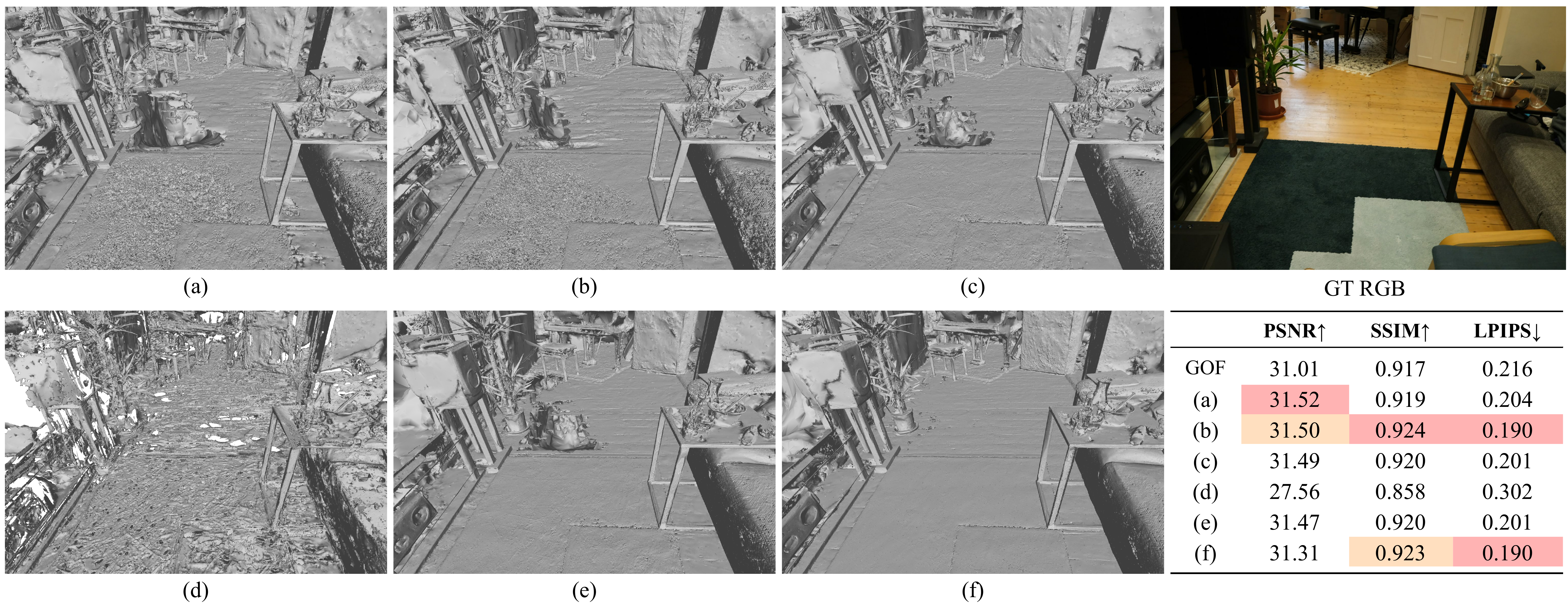}  
  \caption{\label{fig:ablation}%
        Ablation study on the impact of our proposed components on the Mip-NeRF360 `room' scene~\cite{barron2022mipnerf}. We demonstrate the effect of adding our proposed components to the baseline GOF framework~\cite{yu2024gof} on geometry and Novel View Synthesis (NVS).
        The inset table provides quantitative NVS metrics on the test set of the Mip-NeRF360 `room' scene.
        (a) the baseline$+$multview RGB loss; (b) (a) $+$MVS-guided initialization; (c) (a)$+$single-view relative depth loss; (d) (a)$+$multiview relative depth loss$+$mean depth; (e) (a)$+$multiview relative depth loss; (f) (e)$+$MVS-guided initialization$+$multview normal and depth distortion losses (our full method).
        }
\end{figure*}

Relying solely on geometric constraints during optimization can be insufficient, as sparse or inaccurate initial Gaussian placements may lead the optimization process into suboptimal local minima. This issue is particularly pronounced in image regions exhibiting high appearance variation (\Fig{ablation} (e)). Therefore, we introduce a robust MVS-guided initialization scheme to establish a more favorable starting point.

First, we aggregate the filtered multiview depth maps (obtained during preprocessing, see \Sec{overview}) from an MVS method~\cite{schoenberger2016mvs} into a unified 3D point cloud. As the resulting point cloud is typically extremely dense and contains redundancy, we apply multi-voxel grid filtering to achieve a target point count $K'$. Specifically, we construct a voxel grid with an initial voxel size $l$. From the points contained within each non-empty voxel, we randomly sample one point. We then check if the resulting filtered point count $|P|$ is less than or equal to the target $K'$. If so, the filtering terminates. Otherwise, the voxel size is increased $l \leftarrow 1.5l$ and the sampling process is repeated until the point count $ |P|$ is less than or equal to the target $K'$.

Although this filtering removes redundancy, the point cloud $P$ may still contain noisy points originating from MVS estimation errors. To mitigate the impact of these potential outliers, we adapt the opacity-based pruning mechanism inspired by the adaptive density control (ADC) of the original 3DGS~\cite{kerbl20233dgs}. Gaussian splatting optimization exhibits a tendency to rapidly decrease the opacity $\alpha_k$ of Gaussians that do not contribute positively to reconstructing the scene's appearance (i.e., those hindering the reduction of the photometric loss). We leverage this tendency specifically for noise removal during an initial optimization phase. For a predefined number of iterations, $N_{prun}$, we freeze the updates for only Gaussian centers and optimize all parameters except for centers. During this phase, Gaussians whose opacity $\alpha_k$ falls below a specified threshold $\tau$ are removed. Following this initial pruning phase (after $N_{prun}$ iterations), we unfreeze all parameters and proceed with the standard optimization procedure, incorporating the full ADC mechanisms (densification and splitting) from 3DGS as needed.

This MVS-guided initialization procedure provides a much stronger and geometrically informed starting point for the main optimization, resulting in improved final geometry and appearance reconstruction quality.

\begin{figure*}[th]
  \centering

  \includegraphics[width=.95\linewidth]{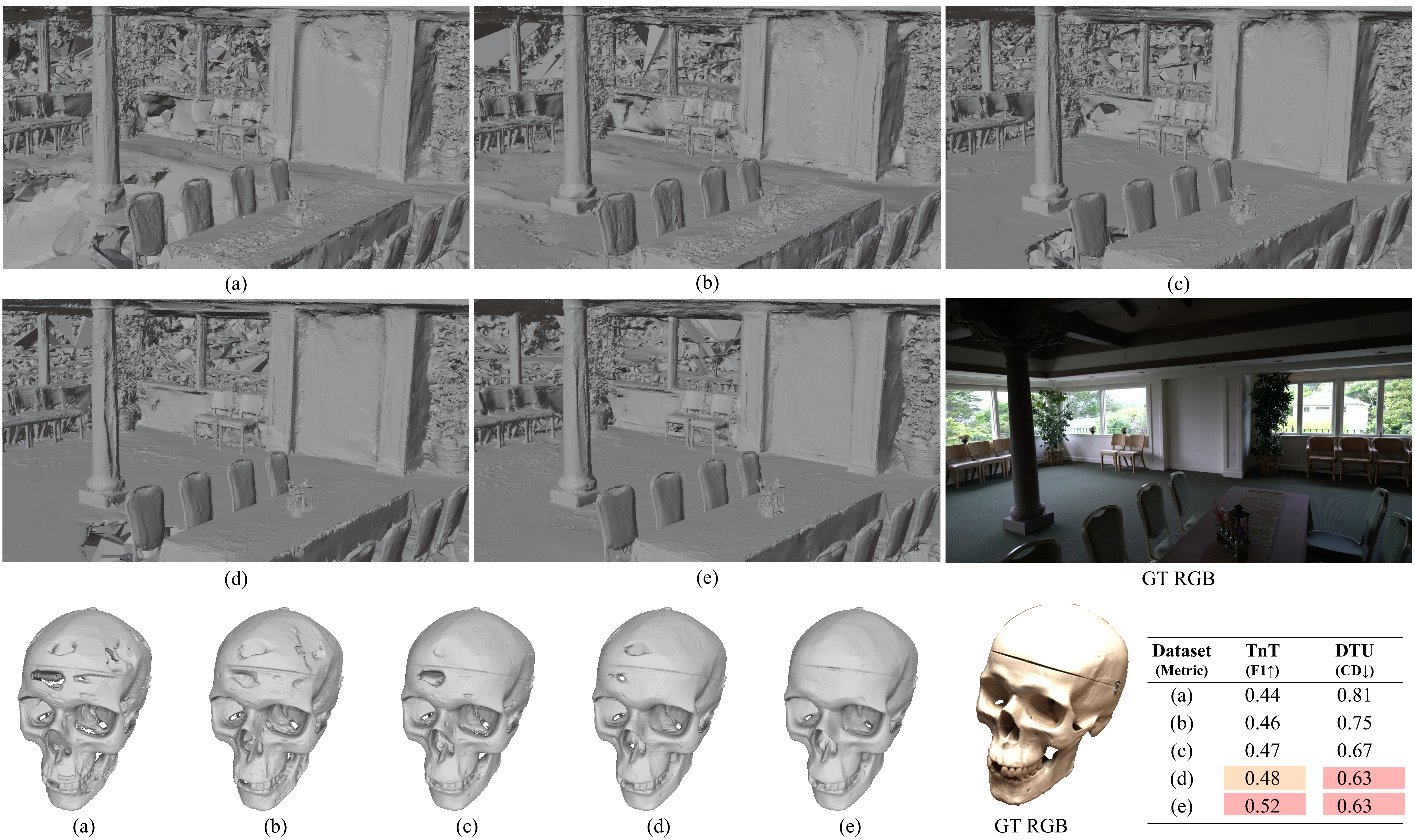}
  
  \caption{\label{fig:ablation2}    
        Ablation study evaluating the impact of our proposed components on the Tanks and Temples (TnT)~\cite{knapitsch2017tanks} and DTU~\cite{jensen2014large} datasets. We start with the GOF framework~\cite{yu2024gof} as a baseline and progressively add our components to demonstrate their effects on geometry reconstruction. The configurations shown are: (a) the baseline $+$multview RGB loss; (b) (a)$+$MVS-guided initialization; (c) (a)$+$single-view relative depth loss; (d) (a)$+$multiview relative median depth loss; (e) (d)$+$MVS-guided initialization$+$multview normal and depth distortion losses (our full method). The inset table reports F1 score and Chamfer Distance (CD) results averaged across all scenes from the DTU and TnT datasets.
        }
\end{figure*}

\begin{figure*}[th]
  \centering

  \includegraphics[width=0.95\linewidth]{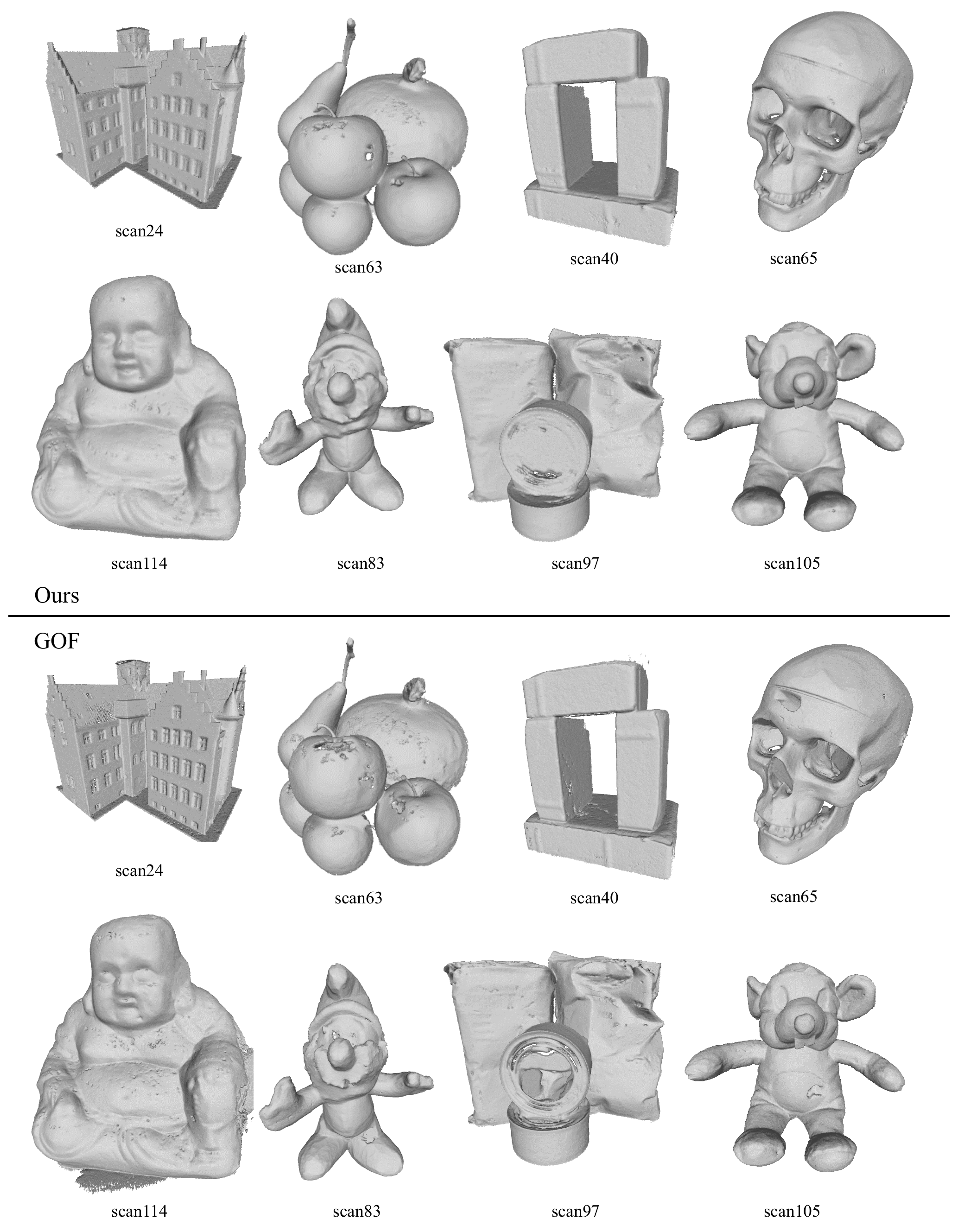}
  
  \caption{\label{fig:comparison_DTU}%
        Visual comparison of 3D meshes reconstructed using GOF~\cite{yu2024gof} and our framework on the DTU dataset~\cite{jensen2014large}.}
\end{figure*}

\begin{figure*}[th]
  \centering
  
  \includegraphics[width=.95\linewidth]{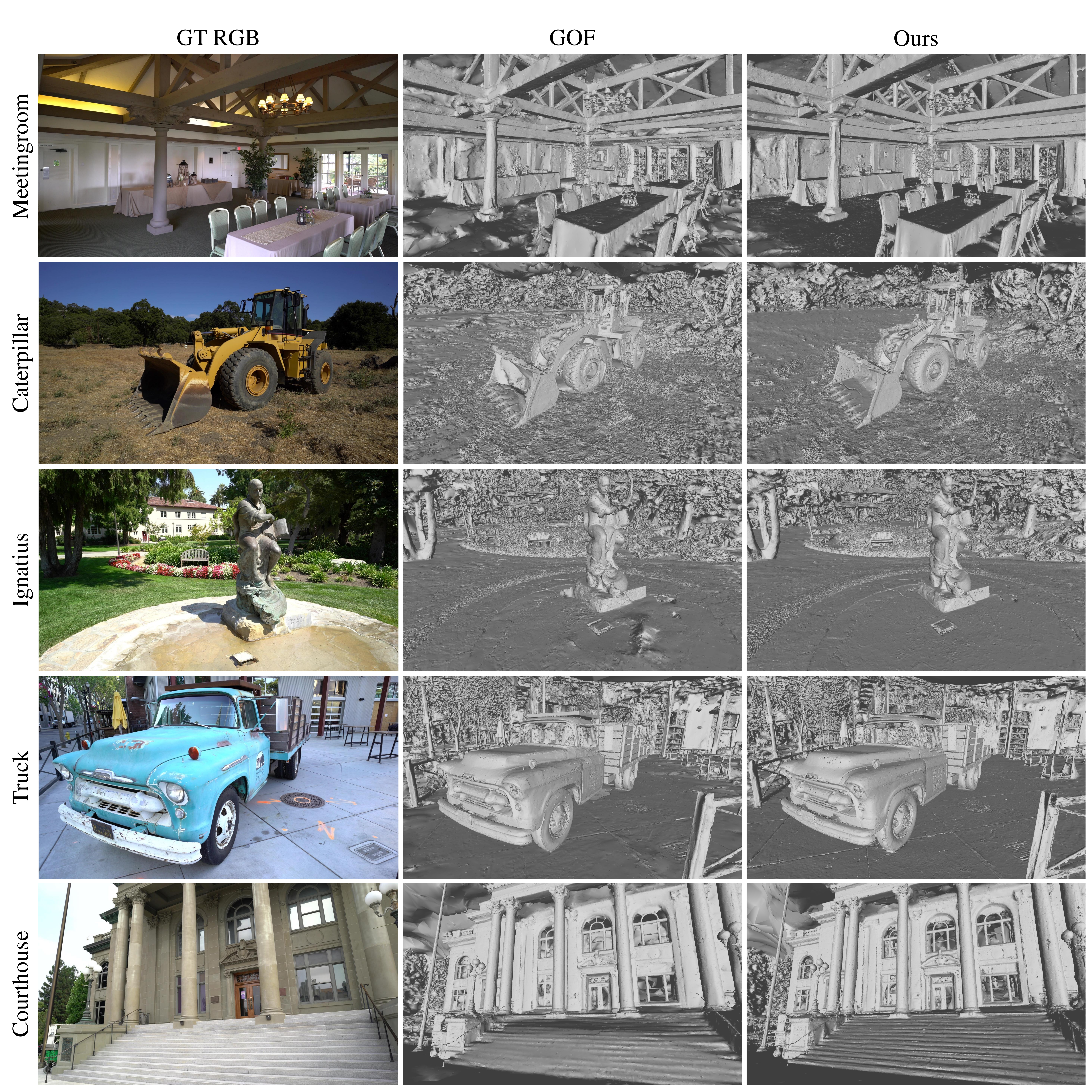}
  
  \caption{\label{fig:comparison_tnt}
  Visual comparison of 3D meshes reconstructed using GOF~\cite{yu2024gof} and our framework on the TnT dataset~\cite{knapitsch2017tanks}.}
\end{figure*}

\begin{table*}[th]
\centering

\resizebox{\textwidth}{!}{

\begin{tabular}{cl|ccccccccccccccc|cc}
 & & 24 & 37 & 40 & 55 & 63 & 65 & 69 & 83 & 97 & 105 & 106 & 110 & 114 & 118 & 122 & Mean & Time \\
 
\hline
\multirow{5}{*}{\rotatebox[origin=c]{90}{implicit}} 
& \raggedleft NeRF~\cite{mildenhall2021nerf} & 1.90 & 1.60 & 1.85 & 0.58 & 2.28 & 1.27 & 1.47 & 1.67 & 2.05 & 1.07 & 0.88 & 2.53 & 1.06 & 1.15 & 0.96 & 1.49 & $>$12h \\

 & \raggedleft VolSDF~\cite{yariv2021volsdf} & 1.14 & 1.26 & 0.81 & 0.49 & 1.25 & 0.70 & 0.72 & 1.29 & 1.18 & 0.70 & 0.66 & 1.08 & 0.42 & 0.61 & 0.55 & 0.86 & $>$12h \\
 
 & \raggedleft NeuS~\cite{wang2021neus} & 1.00 & 1.37 & 0.93 & 0.43 & 1.10 & \cellcolor{orange!25}0.65 & \cellcolor{yellow!20}0.57 & 1.48 & 1.09 & 0.83 & \cellcolor{orange!25}0.52 & 1.20 & \cellcolor{orange!25}0.35 & 0.49 & 0.54 & 0.84 & $>$12h \\

 & \raggedleft NeuS2~\cite{wang2023neus2} & 0.56 & \cellcolor{orange!25}0.76 & 0.49 &\cellcolor{orange!25}0.37 & \cellcolor{yellow!20}0.92 & 0.71 & 0.76 & \cellcolor{orange!25}1.22 & \cellcolor{yellow!20}1.08 & \cellcolor{orange!25}0.63 & 0.59 & \cellcolor{yellow!20}0.89 & 0.40 & \cellcolor{yellow!20}0.48 & 0.55 & \cellcolor{yellow!20}0.70 & \cellcolor{red!30}3.3m \\
 
 & \raggedleft Neuralangelo~\cite{li2023neuralangelo} & \cellcolor{red!30}0.37 & \cellcolor{red!30}0.72 &\cellcolor{orange!25}0.35 & \cellcolor{red!30}0.35 & \cellcolor{red!30}0.87 & \cellcolor{red!30}0.54 & \cellcolor{red!30}0.53 & 1.29 & \cellcolor{red!30}0.97 & 0.73 & \cellcolor{red!30}0.47 & \cellcolor{red!30}0.74 & \cellcolor{red!30}0.32 & \cellcolor{red!30}0.41 & \cellcolor{red!30}0.43 & \cellcolor{red!30}0.61 & $>$12h \\

\hline\hline
\multirow{6}{*}{\rotatebox[origin=c]{90}{explicit}} 
& \raggedleft 3DGS~\cite{kerbl20233dgs} & 2.14 & 1.53 & 2.08 & 1.68 & 3.49 & 2.21 & 1.43 & 2.07 & 2.22 & 1.75 & 1.79 & 2.55 & 1.53 & 1.52 & 1.50 & 1.96 & 11.2 m \\

 & \raggedleft SuGaR~\cite{guedon2024sugar} & 1.47 & 1.33 & 1.13 & 0.61 & 2.25 & 1.71 & 1.15 & 1.63 & 1.62 & 1.07 & 0.79 & 2.45 & 0.98 & 0.88 & 0.79 & 1.33 & $>$ 1h \\
 
 & \raggedleft GaussianSurfels~\cite{dai2024gaussiansurfels} & 0.66 & 0.93 & 0.54 & 0.41 & 1.06 & 1.14 & 0.85 & 1.29 & 1.53 & 0.79 & 0.82 & 1.58 & 0.45 & 0.66 & 0.53 & 0.88 & \cellcolor{orange!25}6.7m \\
 
 & \raggedleft 2DGS~\cite{huang20242dgs} & \cellcolor{yellow!20}0.48 & 0.91 & 0.39 & 0.39 & 1.01 & 0.83 & 0.81 & 1.36 & 1.27 & 0.76 & 0.70 & 1.40 & 0.40 & 0.76 & 0.52 & 0.80 & \cellcolor{yellow!20}10.9m \\
 
 & \raggedleft GOF~\cite{yu2024gof} & 0.50 & 0.82 & \cellcolor{yellow!20}0.37 & \cellcolor{orange!25}0.37 & 1.12 & 0.74 & 0.73 & \cellcolor{red!30}1.18 & 1.29 & \cellcolor{yellow!20}0.68 & 0.77 & 0.90 & 0.42 & 0.66 & \cellcolor{yellow!20}0.49 & 0.74 & 18.4m \\
 
 & \raggedleft Ours & \cellcolor{orange!25}0.43 & \cellcolor{yellow!20}0.79 & \cellcolor{red!30}0.33 & \cellcolor{red!30}0.35 & \cellcolor{orange!25}0.91 & \cellcolor{yellow!20}0.67 & \cellcolor{orange!25}0.55 & \cellcolor{yellow!20}1.28 & \cellcolor{orange!25}1.03 & \cellcolor{red!30}0.61 & \cellcolor{yellow!20}0.56 & \cellcolor{orange!25}0.77 & \cellcolor{orange!25}0.35 & \cellcolor{orange!25}0.44 & \cellcolor{orange!25}0.44 & \cellcolor{orange!25}0.63 & 58.1m + $\alpha$ \\
\end{tabular}
}
\caption{Quantitative comparison of geometry reconstruction on the DTU dataset~\cite{jensen2014large} using Chamfer distance (mm, lower better). The rightmost columns show the mean Chamfer distance across all scenes and average training times (minutes). $\alpha$ is $\sim 4m$, denoting the average time elapsed for estimating MVS depth from RGB images. Results are color-coded by rank: \colorbox{red!30}{1st}, \colorbox{orange!25}{2nd}, and \colorbox{yellow!20}{3rd}.}
\label{tbl:dtu_comparison}
\end{table*}

\begin{table}[ht]
\centering
\resizebox{\columnwidth}{!}{
\begin{tabular}{l c c c c c c c c}
\toprule
 & \multicolumn{3}{c}{\textbf{Implicit}} & \multicolumn{5}{c}{\textbf{Explicit}} \\
\cmidrule(lr){2-4}\cmidrule(lr){5-9}
 & NeuS & Geo-Neus & Neuralangelo & SuGaR & 3DGS & 2DGS & GOF & Ours \\
\midrule
Barn 
  &                   0.29 
  &                   0.33 
  & \cellcolor{red!30}0.70 
  & 0.14 
  & 0.13 
  & 0.41 
  & \cellcolor{yellow!20}0.51 
  & \cellcolor{orange!25}0.55 \\
Caterpillar 
  & 0.29 
  & 0.26 
  & \cellcolor{yellow!20}0.36 
  & 0.16 
  & 0.08 
  & 0.23 
  & \cellcolor{orange!25}0.41 
  & \cellcolor{red!30}0.43 \\
Courthouse 
  & 0.17 
  & 0.12 
  & \cellcolor{orange!25}0.28 
  & 0.08 
  & 0.09 
  & 0.16 
  & \cellcolor{orange!25}0.28
  & \cellcolor{red!30}0.39\\
Ignatius 
  & \cellcolor{orange!25}0.83 
  & \cellcolor{yellow!20}0.72
  & \cellcolor{red!30}0.89 
  & 0.33 
  & 0.04 
  & 0.51 
  & 0.68
  & 0.67\\
Meetingroom 
  & 0.24 
  & 0.20 
  & \cellcolor{orange!25}0.32
  & 0.15 
  & 0.01 
  & 0.17
  & \cellcolor{yellow!20}0.28
  & \cellcolor{red!30}0.39\\
Truck 
  & 0.45 
  & 0.45 
  & \cellcolor{yellow!20}0.48 
  & 0.26 
  & 0.19
  & 0.45 
  & \cellcolor{orange!25}0.59
  & \cellcolor{red!30}0.67\\
\midrule
Mean 
  & 0.38 
  & 0.35 
  & \cellcolor{orange!25}0.50 
  & 0.19
  & 0.09 
  & 0.32 
  & \cellcolor{yellow!20}0.46
  &  \cellcolor{red!30}0.52\\
Time 
  & >24h 
  & >24h 
  & >24h 
  & ~1h 
  & \cellcolor{red!30}14.3m 
  & \cellcolor{orange!25}15.5m 
  & \cellcolor{yellow!20}24.2m
  & 75.1m + $\alpha$\\
\bottomrule
\end{tabular}
}
\caption{Quantitative evaluation of 3D reconstructions on the Tanks and Temples Dataset~\cite{knapitsch2017tanks}. Results show F1 scores and training times for the compared methods. $\alpha$ is $\sim 35m$, denoting the average time elapsed for estimating MVS depth from RGB images. Higher F1 scores indicate better performance. Results are color-coded by rank: \colorbox{red!30}{1st}, \colorbox{orange!25}{2nd}, and \colorbox{yellow!20}{3rd}.
}
\label{tbl:tnt_geo_comparison}
\end{table}

\begin{table}[t]
\centering
\resizebox{\columnwidth}{!}{
\begin{tabular}{l|ccc|ccc}

 & \multicolumn{3}{c|}{Outdoor Scene} & \multicolumn{3}{c}{Indoor scene} \\
 & PSNR $\uparrow$ & SSIM $\uparrow$ & LPIPS $\downarrow$ & PSNR $\uparrow$ & SSIM $\uparrow$ & LPIPS $\downarrow$ \\
\hline
NeRF & 21.46 & 0.458 & 0.515 & 26.84 & 0.790 & 0.370 \\
Deep Blending & 21.54 & 0.524 & 0.364 & 26.40 & 0.844 & 0.261 \\
Instant NGP & 22.90 & 0.566 & 0.371 & 29.15 & 0.880 & 0.216 \\
MERF & 23.19 & 0.616 & 0.343 & 27.80 & 0.855 & 0.271 \\
MipNeRF360 & 24.47 & 0.691 & 0.283 & \cellcolor{red!30}31.72 & 0.917 & \cellcolor{yellow!20}0.180 \\
\hline\hline
Mobile-NeRF & 21.95 & 0.470 & 0.470 & - & - & - \\
BakedSDF & 22.47 & 0.585 & 0.349 & 27.06 & 0.836 & 0.258 \\
SuGaR & 22.93 & 0.629 & 0.356 & 29.43 & 0.906 & 0.225 \\
BOG & 23.94 & 0.680 & 0.263 & 27.71 & 0.873 & 0.227 \\
\hline\hline
3DGS & 24.64 & 0.731 & 0.234 & 30.41 & 0.920 & 0.189 \\
Mip-Splatting & 24.65 & 0.729 & 0.245 & \cellcolor{yellow!20}30.90 & 0.921 & 0.194 \\
2DGS & 24.34 & 0.717 & 0.246 & 30.40 & 0.916 & 0.195 \\
GOF & \cellcolor{yellow!20}24.82 & \cellcolor{orange!25}0.750 & \cellcolor{yellow!20}0.202 & 30.79 & \cellcolor{orange!25}0.924 & 0.184 \\
3DGS-MCMC & \cellcolor{red!25}25.13 & \cellcolor{red!30}0.759 & \cellcolor{red!30}0.194& \cellcolor{orange!25}31.60& \cellcolor{red!30}0.932 & \cellcolor{orange!25}0.174\\
Ours & \cellcolor{orange!30}24.84 & \cellcolor{yellow!20}0.748 & \cellcolor{orange!25}0.197 & 30.65 & \cellcolor{yellow!20}0.922 & \cellcolor{red!30}0.172 \\

\end{tabular}
}
\caption{Quantitative comparison of novel view synthesis quality on the Mip-NeRF360 dataset~\cite{barron2022mipnerf}. Results are color-coded by rank: \colorbox{red!30}{1st}, \colorbox{orange!25}{2nd}, and \colorbox{yellow!20}{3rd}.}
\label{tbl:nvs_comparison}
\end{table}

\section{Experiments}

\subsection{Implementation Details}

We configured the hyperparameters  as follows. The loss weights were set to $\lambda_{rel}=1$ for our relative depth loss, $\lambda_d=100$ for the depth distortion loss, and $\lambda_n=0.05$ for the normal consistency loss. For the MVS-guided initialization, the initial pruning phase duration was set to $N_prun=2000$ iterations, the target point count after voxel filtering was $K'=6$M, and the initial voxel size was $l=0.005$ meter. The thresholding hyperparameter $s$ in \Eq{relative_depth_loss} was initialized to 0.15 and annealed to 0.1 at 7,000 iterations and further reduced to 0.05 at 20,000 iterations. The same hyperparameters are used across all experiments.

\subsection{Analysis}

\paragraph*{Computational Cost}

While the baseline GOF~\cite{yu2024gof} performs volume rendering from a single viewpoint per iteration, our multiview losses require rendering from three viewpoints (corresponding to adjacent view triplets identified during preprocessing). Our method also requires additional computation time for estimating MVS depth from input RGB images. Additionally, our method commences optimization with a denser and more reliable set of initial points due to MVS-guidance and tends to retain a larger number of Gaussians to capture fine geometric details, facilitated by our multiview relative depth loss. This increased Gaussian count leads to a slight increase in optimization time compared to the baseline. We report average computation times on the DTU~\cite{jensen2014large} and Tanks and Temples~\cite{knapitsch2017tanks} datasets in \Tbls{dtu_comparison} and \ref{tbl:tnt_geo_comparison}, respectively.

\paragraph*{ Mean Depth vs. Median Depth}

We evaluate the impact of using either mean depth or median depth for the rendered depth term $\mathbf{D}_r$ within our relative depth loss formulation (\Eq{relative_depth_loss}). As discussed in \Sec{gs_mv_reg}, utilizing the rendered mean depth suffers from several potential caveats: It can yield inaccurate estimates for semi-transparent objects and is sensitive to unstable rendering weights during optimization. Consequently, employing the mean depth within our relative depth loss leads to significantly degraded geometric reconstruction quality and even adversely affects novel view synthesis performance, as demonstrated in \Fig{ablation} (d). In contrast, the median depth-based relative depth loss effectively regularizes the Gaussians, resulting in substantially improved geometric accuracy (\Fig{ablation} (e)).

\paragraph*{Effect of Multiview RGB Loss}

Employing multiview RGB photometric losses ($\mathcal{L}^v_c$) primarily enhances novel view synthesis performance by enforcing appearance consistency across views. However, as shown in \Fig{ablation} (a) and its accompanying table, this component does not improve the accuracy of the geometric reconstruction.

\paragraph*{Effect of Multiview Relative Depth Loss}

Extending our relative depth loss to a multiview formulation ($\mathcal{L}^v_{rel}$) allows the optimization gradient to influence the parameters of more Gaussians simultaneously within a single iteration. This enhances the regularization effect, leading to less noisy and smoother surface reconstructions compared to applying the loss only from a single view, as evidenced in \Fig{ablation} (c, e) and \Fig{ablation2} (c, d).

\paragraph*{Effect of Multiview Normal and Depth Distortion Losses}

As demonstrated by specific regions (e.g., the sofa and floor in \Fig{ablation} (e, f)), the multiview extension of the geometric consistency losses enhances mesh smoothness and fidelity.

\paragraph*{Effect of MVS-guided Initialization}

Owing to our robust MVS-guided initialization (\Sec{gs_mv_reg}), the optimization process starts from a strong geometric foundation. This enables the successful reconstruction of challenging geometry, such as the highly specular floor regions in \Fig{ablation} (e, f), and the upper region of the skull in \Fig{ablation2} (d, e). It is noteworthy that using this initialization procedure alone, without the regularization of our relative depth loss, still results in noisy and inaccurate surface reconstruction (\Fig{ablation} (b) and \Fig{ablation2} (b)), highlighting the synergy between the components of our method.

\subsection{Comparison}

We compare our framework against various state-of-the-art methods employing different scene representations, including implicit functions~\cite{mildenhall2021nerf, yariv2021volsdf, wang2021neus, wang2023neus2, li2023neuralangelo,hedman2018deepblending, muller2022instant, reiser2023merf, barron2022mipnerf}, meshes~\cite{chen2023mobilenerf, yariv2023bakedsdf, guedon2024sugar, reiser2024bog}, and point-based approaches~\cite{kerbl20233dgs, dai2024gaussiansurfels, huang20242dgs, yu2024gof, yu2024mipsplatting}, evaluating performance on both novel view synthesis and 3D surface reconstruction tasks.

\paragraph*{Datasets}

Experiments were conducted on DTU~\cite{jensen2014large}, Tanks and Temples (TnT)~\cite{knapitsch2017tanks}, and Mip-NeRF360~\cite{barron2022mipnerf}. DTU (15 scenes, 49/69 views, 1600x1200) and TnT were used for geometry evaluation, while Mip-NeRF360 was used for novel view synthesis evaluation following the 3DGS protocol~\cite{kerbl20233dgs}.

\paragraph*{Geometry Reconstruction}
We employ surface reconstruction procedures tailored to the scale of the target scenes. For the object-scale scenes within the DTU dataset, we render median depth images from the optimized Gaussians at the training viewpoints, perform TSDF fusion~\cite{curless1996volumetric,Zhou2018} on these depth maps, and finally extract a mesh from the resulting signed distance field using the Marching Cubes algorithm~\cite{lorensen1998marching}. For the large-scale scenes present in the Mip-NeRF360 and Tanks and Temples (TnT) datasets, we first generate an opacity field from the optimized Gaussians and then extract a mesh using the Marching Tetrahedra algorithm, following~\cite{yu2024gof}.

We compare our geometric results against methods based on implicit~\cite{mildenhall2021nerf, yariv2021volsdf, wang2021neus, wang2023neus2, li2023neuralangelo} and explicit~\cite{kerbl20233dgs, guedon2024sugar, dai2024gaussiansurfels, huang20242dgs, yu2024gof} representations. As reported in \Tbl{dtu_comparison} and shown in \Figs{comparison_DTU} and \ref{fig:comparison_tnt}, our method achieves state-of-the-art performance among explicit representation-based approaches, and our results are highly comparable to Neuralangelo~\cite{li2023neuralangelo}, the best performance implicit method. Note that, as shown in \Tbl{tnt_geo_comparison}, our method outperforms Neuralangelo on the large-scale scenes in the TnT dataset.

\paragraph*{Novel View Synthesis}

We evaluate our method for novel view synthesis on the Mip-NeRF360 dataset, comparing it against state-of-the-art approaches that use implicit functions (NeRF~\cite{mildenhall2021nerf}, Deep Blending~\cite{hedman2018deepblending}, Instant NGP~\cite{muller2022instant}, MERF~\cite{reiser2023merf}, MipNeRF360~\cite{barron2022mipnerf}), meshes (Mobile-NeRF~\cite{chen2023mobilenerf}, BakedSDF~\cite{yariv2023bakedsdf}, SuGaR~\cite{guedon2024sugar}, BOG~\cite{reiser2024bog}), and point-based representations (3DGS~\cite{kerbl20233dgs}, Mip-Splatting~\cite{yu2024mipsplatting}, 3DGS-MCMC~\cite{kheradmand20243d}, 2DGS~\cite{huang20242dgs}, GOF~\cite{yu2024gof}). As presented in Table~\ref{tbl:nvs_comparison} and illustrated in Figure~\ref{fig:comp_nerf360_nvs}, our method achieves the second-best overall performance. It is surpassed only by 3DGS-MCMC~\cite{kheradmand20243d}, a method that aims mainly for appearance fidelity, without addressing accurate geometric reconstruction. Our analysis (Figure~\ref{fig:ablation}) indicates that our regularization strategy, designed to promote high-fidelity and smooth geometry, incurs only a minor decrease in novel view synthesis quality. Consequently, our method offers a compelling trade-off, significantly improving geometric quality while largely maintaining the high-fidelity appearance reconstruction characteristic of Gaussian Splatting.

\section{Conclusion}

This paper introduced an effective multiview geometric regularization strategy for Gaussian Splatting to reconstruct geometrically accurate radiance fields. This strategy builds on the key insight that MVS-derived depth points and Gaussian Splatting reconstructions offer complementary geometric information.
Our extensive experiments demonstrate that this proposed regularization scheme significantly improves geometric accuracy while often enhancing novel view synthesis performance compared to baseline methods. Future work could explore the integration of monocular geometric priors (e.g., estimated normals or relative depths) to further improve reconstruction in regions where multiview stereo cues are inherently unreliable or absent.

\paragraph*{Limitations}

Our multiview geometric regularization strategy relies on the fundamental assumption that Multi-View Stereo (MVS) depth estimates are generally more accurate than the initial or unregularized Gaussian Splatting geometry, especially in well-textured regions. Consequently, when MVS depth estimation is unreliable (e.g., due to extreme texturelessness, challenging non-Lambertian surfaces, or hard shadow regions), the effectiveness of our regularization may be limited. A promising direction for future work is to explore the incorporation of image semantics, such as shadow detection and intrinsic image decomposition, into the regularization scheme to address this limitation. Furthermore, similar to most MVS-based and standard Gaussian Splatting approaches, our method struggles with the accurate geometric reconstruction of semi-transparent or transparent objects, as reliable geometric cues are inherently difficult to obtain from either MVS or the splatting representation itself in such cases.

\section*{Acknowledgements}
We thank the anonymous reviewers for their valuable feedback.
This work was supported by 
NRF grants (RS-2023-00280400, RS-2024-00451947) 
and IITP grants (RS-2022-II220290, RS-2024-00437866, RS-2021-II212068) 
funded by the Korean government (MSIT). 

\bibliographystyle{eg-alpha-doi} 
\bibliography{egbibsample}       


\end{document}